\documentclass[12 pt]{article}
\usepackage{cite}
\usepackage{graphicx}
\graphicspath{ {images/} }
\usepackage{amsmath}
\usepackage{amssymb}
\usepackage{graphicx}
\usepackage{subcaption}

\usepackage{geometry}
\pdfoutput=1

\title{\vspace{-3.0cm}\line(1,0){450}\\\textbf{Deep reinforcement learning for robotic manipulation-the state of the art\\\line(1,0){450}}}
\author{\textbf{Smruti Amarjyoti}\\ \footnotesize  Robotics Institute, School of Computer Science,\\\footnotesize Carnegie Mellon University\\\footnotesize samarjyo@andrew.cmu.edu}
\date{\vspace{-5ex}}

\begin{document}
\maketitle
\vspace{5.0mm}\begin{abstract}
The focus of this work is to enumerate the various approaches and algorithms that center around application of reinforcement learning in robotic manipulation tasks. Earlier methods utilized specialized policy representations and human demonstrations to constrict the policy. Such methods worked well with continuous state and policy space of robots but failed to come up with generalized policies. Subsequently, high dimensional non-linear function approximators like neural networks have been used to learn policies from scratch. Several novel and recent approaches have also embedded control policy with efficient perceptual representation using deep learning. This has led to the emergence of a new branch of dynamic robot control system called deep reinforcement learning(DRL). This work embodies a survey of the most recent algorithms, architectures and their implementations in simulations and real world robotic platforms. The gamut of DRL architectures are partitioned into two different branches namely, discrete action space algorithms(DAS) and continuous action space algorithms(CAS). Further, the CAS algorithms are divided into stochastic continuous action space(SCAS) and deterministic continuous action space(DCAS) algorithms. Along with elucidating an organisation of the DRL algorithms this work also manifests some of the state of the art applications of these approaches in robotic manipulation tasks.
\end{abstract}

\section{Introduction}

Reinforcement learning (RL) provides a neuropsychological and cognitive science perspective to animal behavior and sequential decision making. Recent studies in cognitive science have also demonstrated analogies between the dopaminergic neurons in brains and temporal difference (TD) reinforcement learning algorithms. Other than the nature derived inspiration, several successful implementations of reinforcement learning (RL) in controlling dynamic robotic systems for manipulation, locomotion and autonomous driving \cite{koberba}, \cite{dise}, \cite{kober} have proven the previously theoretical concept to be applicable in real time control of physical systems. Many of these methods use specialized policy structures to represent policies in order to put a cap on the number of iterations that are needed for optimizing the behaviour. Though efficient there is a loss of generality in adopting such an approach as it constricts the policy space to some specific trajectories \cite{gps}. Thus, non-linear function approximators like neural networks are used to parametrize the policy. This removes the requirement of using hand engineered policy representations and human supplied demonstrations to initialized them. Moreover, the use of higher number of parameters also theoretically ensures learning of complex behaviours that wouldn't have been possible with linear man made policies. \\
Another important development in the field of RL has been indirectly borrowed from enormous successes of deep convolutional neural networks(CNN) \cite{alex} in image feature extraction. A direct implication of CNNs in reinforcement learning was the use of image pixels as states instead of joint parameters, which was widely in practice in RL landscape. Use of such an expressive parametrization also enabled learning of value function and policies that were previously deemed complicated. The paper by Riedmiller \cite{nfqa} demonstrated that neural networks can effectively be used as q-function approximators using neural fitted q-iteration algorithm. Later introduction of convolutional networks by Mnih et al. \cite{mnih} turned neural networks based q learning as a base for DRL. Some of the ideas that were introduced like mini batch training and concept of target networks were pivotal to the success of non-linear RL methods. But, the initial algorithms were used to play classic Atari 2600 games with pixels as inputs and discrete actions as policy. The result were extraordinary with the artificial agent getting scores that were higher than human level performance and other model based learning methods. Attempts have been made to use deep q-learning (DQN) for high dimensional robotics tasks but with a very little success \cite{bax}. This is essentially because of the fact that most of the physical control tasks have high dimensional action spaces with continuous real valued action values. This posed a problem for introducing DQNs in manipulation tasks as they act as value function approximators and an additional iterative optimization process is necessary to use them for continuous spaces. The algorithms falling under this class are categorized into a group called discrete action space algorithms(DAS) as they are efficient only in discrete action domains.  \\
Another approach to parametrization of RL policies is to encode  the policy directly and search for optimal solutions in the policy space. These methods known as policy search methods are popular as it gives an end-to-end connection between the states and feasible actions in the agent environment. The parameters can then be perturbed in order to optimize the performance output \cite{book}. The advantage of this process over the earlier value approximation methods is that the policies are integrated over both action and state space, thus the search is more comprehensive than Q-learning. And it also solves the discrete action problem as the output policy, $\pi(a|s)$ is a stochastic distribution over the action given a particular state. Thus, the policy representation provides probabilities over over action in a continuous space \cite{trpo}. This class of continuous action algorithms are grouped into continuous action space algorithms(CAS). They include policy gradient and policy iteration algorithms that encode the policy directly and search over entire policy space. Initially developed and experimented on low dimensional state spaces, CAS algorithms have been integrated into CNN architecture in algorithms like deep deterministic policy gradients (DDPG) \cite{ddpg}.  \\
The CAS RL algorithms can further be divided into two subcategories, stochastic continuous action space(SCAS) and deterministic continuous action space(DCAS) algorithms. The main difference between both of the methods is basically the sample complexity. Even though stochastic policy gradient methods provide a better coverage of the policy search space, they require a large number of training samples in order to learn the policy effectively \cite{trpo}. This is quite infeasible in robotic applications as exploration and policy evaluation comes at a price in such domains. Several methods like natural policy gradients and trust region policy gradients were developed in order to make policy search effective by adding additional constraints on the search process to restrict the agent to explore only promising regions. But, the discovery of deterministic policy gradients has led to an easier method whose performance surpasses stochastic policy algorithms as proven empirically by Silver et al \cite{dpg}. \\
The most important contribution of this paper is the organisation of the assortment of DRL algorithms on the basis of their treatment of action spaces and policy representations. Present DRL methodologies in literature are classified into the groups, DAS, CAS, SCAS and DCAS whose details has been described above. Following sections include a background of the evolution of reinforcement learning and preliminaries laying a foundation to understand the algorithms and description of some of the basic algorithms encompassing DRL. Experiments and real time implementations associated with these methods are also described to give an insight into the practical complexity of implementing these algorithms on physical robots/simulations.\\

\section{Background}

\subsection{Preliminaries}
All of the reinforcement learning methods studied in this paper are basically control problems in which an agent has to act in a stochastic environment by choosing action in a sequential manner over several time steps, with an intention to maximise the cumulative reward. The problem is modelled as a \textit{Markov decision process} (MDP) which comprises of a state space $S$, an action space $A$, an initial state distribution with density $p_1(s_1)$, a stationary transition dynamics model with density $p(s_{t+1}|s_t,a_t)$ that satisfies the Markov property $p(s_{t+1}|s_1,a_1,....s_t,a_t)=p(s_{t+1}|s_t,a_t$ for any trajectory in the state-action space and a reward function $R(s_t,a_t):S \times A \rightarrow \mathbb{R}$. A policy can be defined as the mapping of state to action distributions and the objective of an MDP is to find the optimal policy. Generally a policy is stochastic and is denoted by $\pi_\theta: S \rightarrow P(A)$, where $P(A)$ is the probability distribution of performing that action and $\theta \in \mathbb{R}^n$ is a vector of parameters that define the policy, $\pi_\theta(a_t,s_t)$. A deterministic policy on the other hand is denoted by $\mu(s_t)$ and is a discrete mapping of $S \rightarrow A$.\\
A agent uses the policy to explore the environment and generate trajectories of states, rewards and actions, $\zeta_{1:t}=s_1,a_1,r_1,....,s_t,a_t,r_t$. The total return or performance is determined by calculating the total discounted reward from time step $t$ onwards. 
\begin{equation}
R_t^\gamma = \sum_{k=t}^{\inf} \gamma^{k-t}R(s_k,a_k) \ ,where \  0<\gamma<1
\end{equation}
Value function of a particular state is defined as the expected total discounted reward if an agent were to initiate from that particular state and generate trajectories thereafter.
\begin{equation}
V^{\pi}(s)=\mathbb{E}[R^{\gamma}|S=s;\pi]
\end{equation} 
The action-value function on the other hand is defined as the expected discounted reward if the agent takes an action $a$ from a state $s$ and follows the policy distribution thereafter.
\begin{equation}
Q^{\pi}(s,a)=\mathbb{E}[R^{\gamma}|S=s;A=a;\pi]
\end{equation} 
The agent's overall goal is to obtain a policy that results in maximisation of cumulative discounted reward form the start state. This is denoted by finding appropriate $\pi$ for the performance objective $J(\pi)=\mathbb{E}[R^{\gamma}|\pi]$.\\
The density of the state $s^{'}$ after transitioning for t time steps from initial state $s$ is given by $p(s {\rightarrow} s^{'},t,\pi)$. The discounted state distribution is then given by $\rho^{\pi}(s^{'})=\int_s \sum_{t=1}^{\inf} \gamma^{t-1}p_1(s)p(s \rightarrow s^{'},t,\pi)ds$. The performance objective can be represented as a unified expectation function,
\begin{equation}
\begin{aligned}
J(\pi_{\theta})=\int_S \rho^{\pi}(s) \int_A \pi_{\theta}(s,a)R(s,a)dads\\
=\mathbb{E}_{s \in \rho^{\pi}, a \in \pi{\theta}} [R(s,a)]
\end{aligned}
\end{equation}

\subsection{Evolution of RL}
Early reinforcement learning(RL) algorithms for prediction and control were focused on the process of refinement of optimal policy evaluation techniques and reduction of computational complexity of the approaches. This led to the emergence of exploration vs exploitation techniques, on-policy and off-policy approaches, model free and model based and various PAC(Probable approximate correct) methods. Although the algorithms were feasible computationally and showed convergence to optimal policies in polynomial time, they posed a major hindrance when applied to generate policies for high dimensional control scenarios like robotic manipulation. Two techniques stand out from the newly developed RL methodologies, namely function approximation and policy search. The philosophy of these two approaches is to parameterize the action-value function and the policy function. Further, gradient of the policy value is taken to search for the optimal policy that results in a global maxima of expected rewards. Moreover, due to the hyper dimensional state-space and continuous action-space the robot operates in, policy search methods are the most viable and possible the only method considered suitable for robotics control.

\subsection{RL for motor control}
Application of RL in robotics have included locomotion, manipulation and autonomous vehicle control \cite{koberba}. Most of the real world tasks are considered episodic and it is also hard to specify a concise reward function for a robotic task. This problem is tackled by the use of a technique called learning by demonstration or apprenticeship learning \cite{IRL}. One of the methods to solve the uncertain reward problem is inverse reinforcement learning where the reward function is updated continuously and an appropriate policy is generated in the end Another effective method to model the policies is the use of motor policies to represent stochastic policy \( \pi(a_t|s_t,t)\), that is inspired from the works of Kober and Peters \cite{kober}. They devised an Expectation Maximization (EM) based algorithm called Policy Learning by Weighing Exploration with Returns(PoWER). When learning motor primitives, they turn this deterministic mean policy into a stochastic policy using additive exploration in order to make model-free reinforcement learning possible. Here, the motor primitives are derived from the concept of Dynamic Motor Primitives (DMPs) that describe movement as a set of differential equations such that any external perturbation is accommodated without losing the motion pattern. Certain other approaches like guided policy search \cite{gpsnn} also introduced more versatile policy representations like differential dynamic programming (DDP). These policy have been used for generating guiding samples to speed up the learning process in non linear policies. This gives more flexibility and generalization than earlier structured policy approaches. But, even though such hybrid model based and specialized policy methods work well in robots, there has always been an interest towards learning policies end-to-end from visual stimulus. Thus, convolutional architectures have been introduced into the domain of RL and motor control, known as visuo-motor control policy networks.

\subsection{RL for visuo-motor control}
Many of the RL methods demonstrated on physical robotic systems have used relatively low-dimensional policy representations, typically with under one hundred parameters, due to the difficulty of efficiently optimizing high-dimensional policy parameter vectors. But the paper by Mnih et al. introduced an effective approach to combine larger policy parameterizations by combining deep learning and reinforcement learning \cite{mnih}. This concept of generating efficient non-linear representations is transferred into robotic tasks of grasping and continuous servoing in some recent research carried out by Levine et al.\cite{levin} and Kober et al \cite{kober2009}.
End-to-end learning of visuo-motor policies is made possible with such an approach which in turn learns the features form the observations that are relevant for the specific task. One of the problems that was encountered with neural network learning of policies was the convergence of some weights to infinity when trained with similar instances of input observations \cite{levineend}. Solving of this difficulty using experience replay methods constituting randomization of the episodes gave the necessary boost to RL in real life control problems. The current state of the art in deep-reinforcement learning includes the algorithms employed by google deepmind research namely DQN (Deep Q network) for discrete actions and Deep deterministic policy gradients (DDPG) for continuous action spaces \cite{Gu}. DQN is a simple value approximation method while DDPG uses a underlying actor-critic framework for policy evaluation. Efficacy of both of these methods have been demonstrated empirically for performing complex robotic manipulation tasks like door opening and ball catching \cite{ddpg}.

\section{DRL topology}
\begin{figure}[h]
\centering
\includegraphics[width=0.65\textwidth]{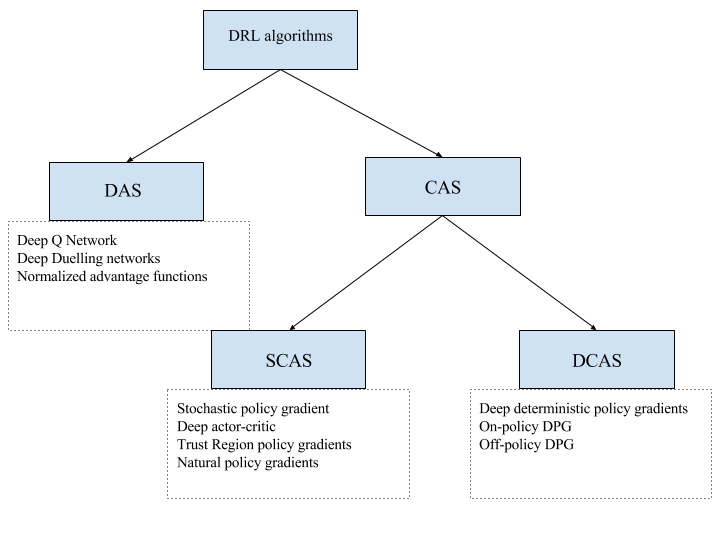}
\caption{The DRL topology.}
\end{figure}
The Deep reinforcement learning algorithms prevailing currently are structured according to the topology in Fig. 1. The initial problem of planning in continuous and high dimensional state spaces can be considered solved due to the extensive use of neural networks with large number of parameters for function/policy modelling. But, the problem at hand now is the mapping of these continuous states to high-dimensional continuous action spaces. The present concentration in the DRL community based on this issue and hence, it seems quite apt to organise the various learning approaches based on this ground. Moreover it also demonstrates the capabilities and limitations of the prevalent algorithms quite clearly.\\
The methods are divided into two sections namely, Discrete action space(DAS) approaches and Continuous action space(CAS) approaches. Further, CAS methods are divided into Stochastic continuous action space(SCAS) and Deterministic continuous action space(DCAS) methods. The various algorithms that come under the purview of DAS are deep q-networks, duelling networks, normalized advantage function and related function approximation approaches to decision making. CAS mostly include policy search approaches that parametrize the policy directly and optimized it using evaluation and gradient based approaches. CAS is further branched into SCAS methods where CNN are used to estimate a stochastic policy and DCAS methods which predicts deterministic policies. Even though this demarcation provides a near comprehensive description of the DRL methods, it misses out on several other RL approaches like likelihood ratio maximisation, black box methods, model-based methods which are not directly related to DRL.
\section{Discrete action space algorithms (DAS)}
\subsection{Deep Q-network (DQN)}
\begin{figure}[h]
\centering
\includegraphics[width=0.45\textwidth]{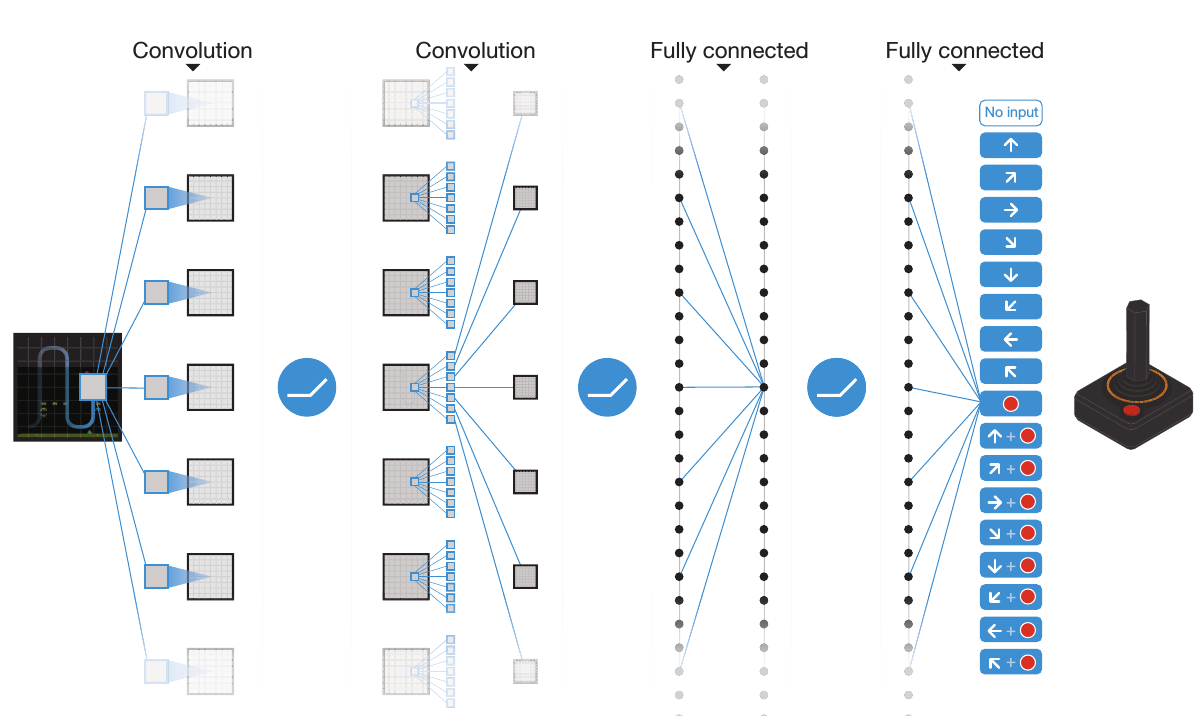}
\caption{Deep-Q-network architecture.}
\end{figure}
The DQN architecture was the first successful integration of deep learning with Q-learning framework \cite{mnih}. Q-learning forms the base of most of the model-free RL algorithms. It includes exploration of the environment using a behaviour policy and learn the Q-function for the possible action-state pairs using the experience that is gathered from the exploration. The following equation described Q-learning, where $\alpha$ is the learning rate and the observations that are obtained after exploration be $a,r,s^{'}$, where $a$ is the action taken, $r$ is the rewards received and $s^{'}$ is the next observed state. 
\begin{equation}
\begin{aligned}
Q(s,a)_{sample}=r(s,a) + \gamma.max_a Q(s^{'},a)\\
Q(s,a)=\alpha Q(s,a)_{sample} + (1-\alpha)Q(s,a)
\end{aligned}
\end{equation}
The only difference between naive Q-learning and DQN is the use of CNN as function approximators instead of linear approximators. The use of hierarchical networks enables the use of continuous high dimensional images as states which estimates the optimal action-value function. RL was considered to be unstable when using non-linear approximators such as a neural network, which is because of the correlations present in the sequence of observations and the correlations between the action-values and the target values. In order to solve this, Mnih et al. devised a method of asynchronous training of the Q-network called experience replay. Here, the experience $e=\{s_t,a_t,r_t,s_{t+1}\}$ is stored in a pool and mini-batches of experiences are accessed during training uniformly. This is then used to optimize the loss function,
\begin{equation}
L(\theta)=\mathbb{E}_{(s,a,r,s^{'})}[(r+\gamma max_aQ(s^{'},a|\theta^{-})-Q(s,a|\theta)^2]
\end{equation}
Fig.2 describes the architecture of the Q-network that consists of 3 convolutional layers, with filter sizes 32x8x8;stride 4, 64x4x4;stride 4 and 64x3x3;stride 2. The final two layers are fully connected layers with 512 neurons and outputs are discrete in the number of actions considered. The activations chosen are rectified linear units. The second important contribution of DQN other than replay buffer was the use of target networks $\hat{Q}(s,a)$ for generating target values for the network's loss function. This helps to reduce oscillations during training and leads to easy convergence. The target network is updated with the online Q network after a specific number of time steps.\\
Execution of this methods is limited to agents requiring discrete action space but, some early works have embedded the DQN technique to learn optimal actions from visual inputs. Zhang et al. have utilized the exact same architecture to learn the optimal control policies for a baxter robot arm.
\begin{figure}
\centering
\begin{subfigure}{.4\textwidth}
  \centering
  \includegraphics[width=.5\textwidth]{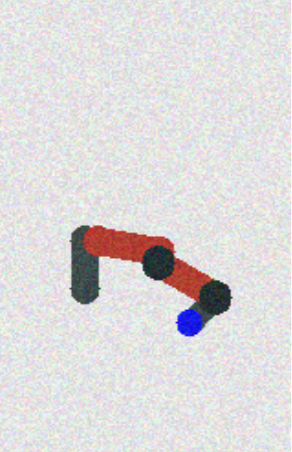}
  \caption{2D baxter arm simulator}
\end{subfigure}
\begin{subfigure}{.4\textwidth}
  \centering
  \includegraphics[width=\textwidth]{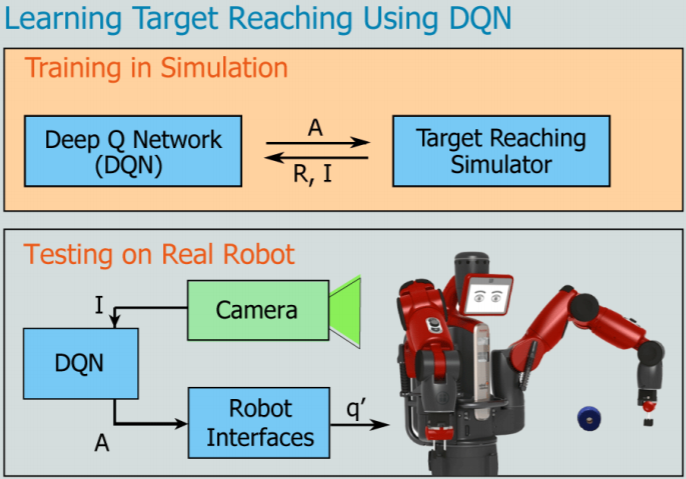}
  \caption{System architecture for applying DQN in robot control.}
\end{subfigure}
\caption{Visuo-motor learning using DQN}
\end{figure}
Instead of controlling the entire 7DOF of the robot arm, only the 4DOF shown in the Fig.3(a) simulation are controlled. The actions are discretized into nine distinct outputs, that include going up, going down or staying put in denominations of 0.02 rad. After training, the network was used to control a real robotic arm with marginal success as it was prone to discrepancies in the input image. Moreover, training in simulation and transferring the control system to real-time robots proved to be detrimental for safety and performance.\\

\subsection{Double Deep Q-networks}
Double deep Q-networks are an improved version of DQN that was first introduced by Hasselet et al. \cite{ddqn}. In Q-learning and DQN, the max operator utilizes the same values for both behaviour policy and evaluation of actions. This in turn gave overestimating value estimates. In order to mitigate this, DDQN uses the target as:
\begin{equation}
y^{DDQN}=r+\gamma Q(s^{'},a^{'}|\theta)|\theta^{-})
\end{equation}

\subsection{Deep duelling network architecture}
\begin{figure}[h]
\centering
\includegraphics[width=0.45\textwidth]{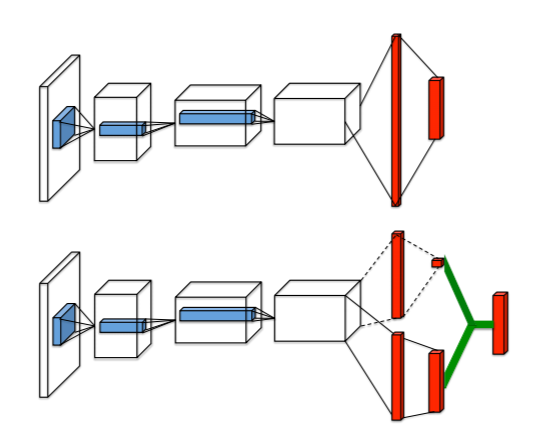}
\caption{The top network is a single stream DQN and the bottom network has dual streams to estimate value and advantage function}
\end{figure}
Duelling architecture, is a model-free algorithm developed by Wang et al. \cite{duel} draws its inspiration from residual RL and the concept of Advantage learning and updating by Baird \cite{baird}. In advantage learning instead of estimation of action-value function, an advantage function is calculated which is defined as the rate of increase of reinforcement when a particular action is taken. The prime importance of advance learning is that the advantage values have higher variance that leads to easy convergence. Moreover, the policy doesn't change discontinuously with changing values. The duelling architecture maintains both $V(s)$ and $A(s,a)$ with a single deep model and a simple output operation combines both these output to get back the $Q(s,a)$ value. As the output is same as DDQN and DQN, this network can be trained with any value iteration method.\\
Considering the duelling network described in Fig. 4 where one stream outputs $V(s|\theta, \beta)$ and other $A(s,a|\theta,\alpha)$. $\theta$ and $\alpha$ denote the convolutional network parameters. The last module is implemented using a forward mapping function:
\begin{equation}
Q(s,a|\theta, \alpha, \beta)=V(s|\theta, \beta) + (A(s,a|\theta, \alpha)-\frac{1}{\|A\|}\sum_a A(s,a|\theta, \alpha) )
\end{equation}
The architecture was used to train an artificial agent learn the 57 games in Atari arcade learning environment from raw pixel observations. The final acquired rewards were compared with that of human performance and DQN networks. Duelling networks performed 75\% better than the naive Q-networks as reported in the paper \cite{duel}. For applications with discrete action robotic tasks duelling architecture can be used for better performance, though a concrete application is missing from literature.

\section{Continuous action space algorithms (CAS)}
\subsection{Normalized advantage functions (NAF)}
Gu et al. proposed a model free approach that used Q-learning to plan in continuous action spaces with deep neural networks, which they refer as Normalized advance functions (NAF). The idea behind NAF is to describe Q function in a way such that its maximum, $argmax_aQ(s_t,a_t)$ can be obtained easily and analytically during the Q-learning update. The inherent processes are equivalent to that of Duelling networks as a separate value function $V(s|\theta)$ and advantage term are estimated. But, the difference is that the advantage in this case is parametrized as a quadratic function of non-linear features of the state:
\begin{equation}
Q(s,a|\theta)=A(s,a|\theta)+V(s|\theta)
\end{equation}
\begin{equation}
A(s,a|\theta)=-\frac{1}{2}(a-\mu(s|\theta)P(s|\theta)(a-\mu(s|\theta)
\end{equation}
$P(s|\theta)$ is a state-dependent, positive definite square matrix that is parametrized by $L(s|\theta)L(s|\theta)^{T}$, where $L$ is a lower triangular matrix whose entries come from the linear activations of the neural network. The rest of the network architecture is similar to that of DQN by Mnih et al. The paper also explored the use of a hybrid model based method by generating imagination rollouts from fitted dynamics model. This incorporated the inclusion of synthetic experience data from the fitted local linear feedback controllers and including them in the replay buffer of on-policy exploration of Q-learning.\\
\begin{figure}[h]
\centering
\includegraphics[width=0.45\textwidth]{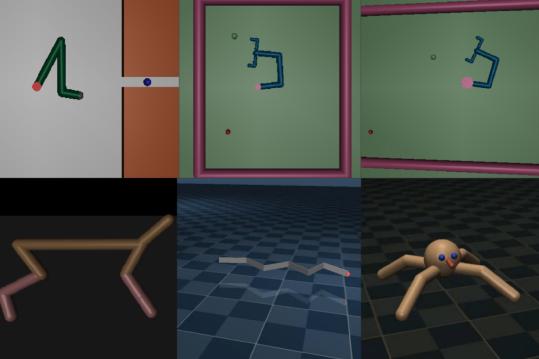}
\caption{Robotic manipulation and locomotion environments in MuJoCo simulator.}
\end{figure}
The algorithms was tested in several robotic environments as shown in Fig.5. The environments include the MuJoCo simulator tasks from Todorov et al. that include 3DOF robotic manipulation tasks where an arm gets reward based on the distance between the end effector and the object to be grasped. It also includes a sex joint 2D swimmer and a four legged ant. Policies learnt with this method showed more precise completion of tasks as compared to deep policy gradient methods \cite{naf}. 

\subsection{Stochastic policy gradient}
Stochastic policy gradient methods parametrize the policy directly rather than trying to optimize the value functions. These are one of the most popular class of continuous action RL algorithms. The central idea behind these algorithms is to adjust the parameters $\theta$ of the policy in the direction of the gradient of the performance, i.e $\Delta_{\theta}J(\pi_{\theta})$. The fundamental theorem underpinning these algorithms is the $policy gradient theorm$ \cite{pg}.
\begin{equation}
\begin{aligned}
\Delta_{\theta}J(\pi_{\theta})=\int_S \rho^{\pi}(s) \int_A \Delta_{\theta} \pi_{\theta}(a|s)Q(s,a)dads\\
=\mathbb{E}[\Delta_{\theta} log \pi_{\theta}(a|s)Q(s,a)]
\end{aligned}
\end{equation} 

The interesting aspect of this theorem is that even though the state distribution depends on the network parameters, the policy gradient doesn't depend on the gradient of the distribution. But, one of the issues that these algorithms have to address is the estimation of the $Q(s,a)$ function as evident from the above equation. Even though policy gradient algorithms provide an end-to-end method for policy search, it is rarely used in robot policy optimization tasks. This is because of the high sample complexity of such algorithms. Policy gradients use on-policy exploration policy and that results in it needing a large number of training data to learn, that is infeasible for robots. 
\begin{figure}[h]
\centering
\includegraphics[width=\textwidth]{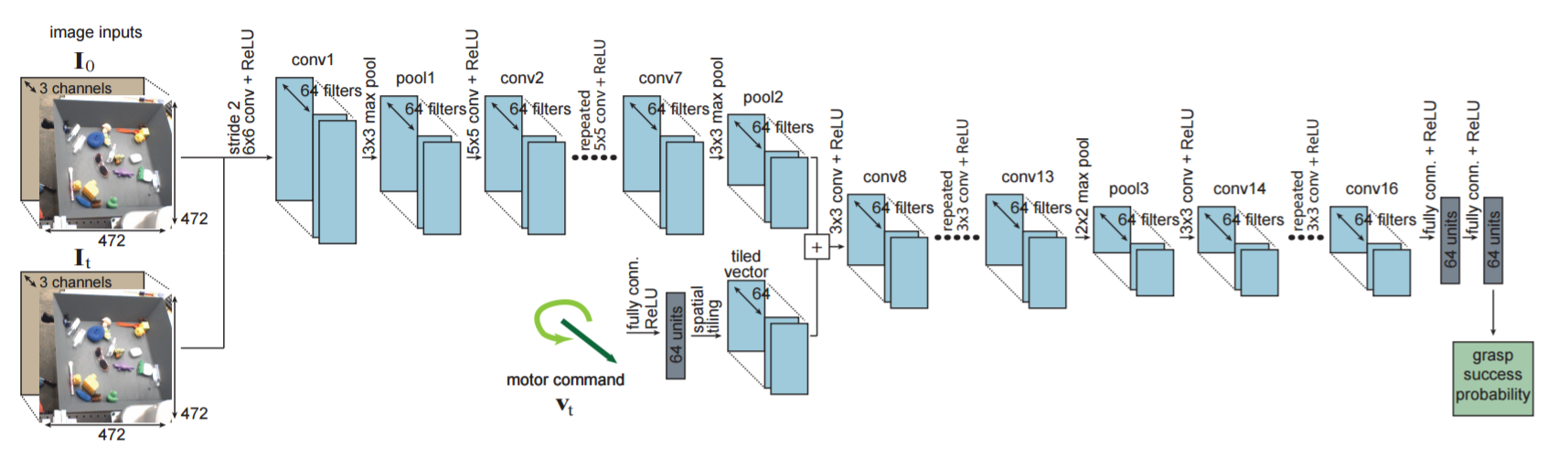}
\caption{The architecture of CNN grasp predictor used in Levine et al. There are a total of 17 conv layers with inputs as the stationary image of object cluttered environment and the online image, the joint configuration is also passed on after the 7th layer.}
\end{figure}
Above figure depicts a stochastic policy gradient algorithm that is used by Levine et al. \cite{levin} for autonomous grasping of objects in cluttered environments. The input is a monocular image showing objects and robot end effector and the robot actions in the 7th layer of the deep network. The output is the probability distribution of the action given the particular state. The network takes 800,000 labelled images to train which gives a clear indication of the sample complexity of SCAS methods. 

\subsection{Stochastic actor-critic methods}
Actor-critic methods are widely used architectures that are again based on the Policy gradient theorem. As depicted from the policy gradient equation, the term $ Q(s,a) $ is missing from the gradient and needs to be calculated. Hence, the critic network estimates this $ Q(s,a) $ value in order to find the derivatives of the actor network,$ \Delta \pi_{\theta}(s) $.

\subsection{Trust Region Policy Optimization}
TRPO is a policy optimization algorithm that restricts the search space of the policy by applying constraints on the output policy distributions. This is done by penalizing the network parameters using a KL divergence loss function upon the parameters, $D^{max}_{KL}(\theta_{old},\theta)$. Intuitively this constraint doesn't let large scale changes to occur in the policy distribution and hence, helps in early convergence of the network.
\begin{figure}[h]
\centering
\includegraphics[width=0.5\textwidth]{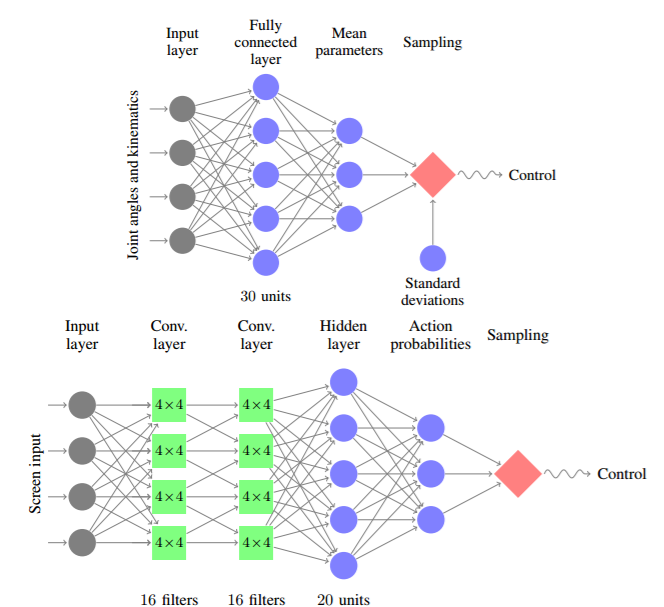}
\caption{Networks used to represent policy in TRPO.}
\end{figure}
The above figure depicts the networks that were used to control the swimmer and hopper tasks in MuJoCo environments. The input state space consisted of joint angles and robot kinematics and the rewards were linear functions.

\subsection{Deterministic policy gradient algorithm}
The deterministic policy gradient algorithm (DPG) is derived from its counterpart stochastic policy gradient algorithm and is dependent of a similar deterministic policy gradient theorem . In continuous action spaces, greedy policy improvement becomes problematic and needs global optimization during policy improvement step \cite{dpg}. As a result the it is more computationally tractable to update the policy parameters in the direction of the gradient of the Q function.
\begin{equation}
\theta^{k+1}=\theta^k+\alpha \mathbb{E}[\Delta_{\theta}Q(s,\mu_{\theta}(s)]
\end{equation} 
Here, $\mu_{\theta}(s)$ is the deterministic policy, $\alpha$ is the learning rate and $\theta$ are the policy parameters. Chain can be applied to the above equation in order to get the policy gradient equation:
\begin{equation}
\theta^{k+1}=\theta^k+\alpha \mathbb{E}[\Delta_{\theta}\mu_{\theta}(s)\Delta_{a}Q(s,\mu_{\theta}(s)]
\end{equation} 

The above update rule can be incorporated into a deep neural network architecture where the policy parameters are updated using stochastic gradient ascent. To realise this an actor-critic method is necessary. The critic estimates the action-value function while the actor derives its gradients from the critic to update its parameters. The gradient of policy parameters is the product of the gradient of Q value with respect to action and the action with respect to the policy parameters. Fig. 8 shows the deterministic actor critic network.
\begin{figure}[h]
\centering
\includegraphics[width=0.5\textwidth]{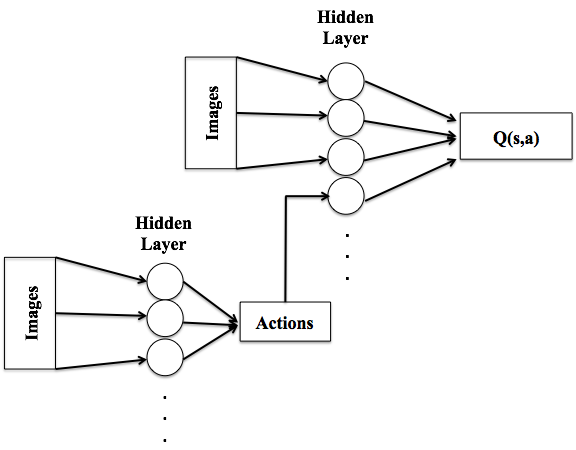}
\caption{Deep deterministic policy gradient network.}
\end{figure}

This is also the basis of DDPG (deep deterministic policy gradient algorithm) which performs better than any other continuous action algorithm. Methods such as NAF and DDPG have been used for learning complex robotic manipulation tasks in real time \cite{Gu}. The authors trained a 7DOF Jaco arm for reaching and door opening tasks without any policy initializations and demonstrations. They used deep network architectures with a 20 dimensional state space consisting of the joint angles, velocities and the end effector pose. The reward function for the reaching task was the distance between the end effector and the object, whereas for door opening the reward was the sum of distance to the door knob and the degrees of rotation of the knob. Another significant contribution of this paper was the use of asynchronous leaning by parallelizing the data collection process. It was proved that using multiple robots reduces the training times by a factor of the number of robots.

\begin{figure}
\centering
\begin{subfigure}{.4\textwidth}
  \centering
  \includegraphics[width=.8\textwidth]{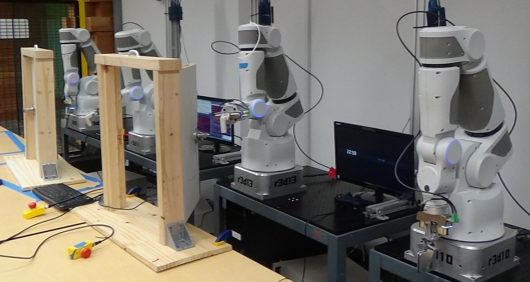}
  \caption{Real-world door opening task setup.}
\end{subfigure}
\begin{subfigure}{.4\textwidth}
  \centering
  \includegraphics[width=\textwidth]{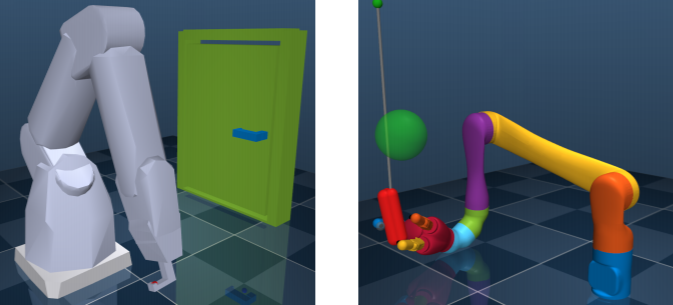}
  \caption{Simulation setup for door opening and object reaching task.}
\end{subfigure}
\caption{Task representation for DDPG and NAF.}
\end{figure}

\section{Discussion}
Algorithmic ideas, theories and implementation details of several deep reinforcement learning algorithms have been delineated in detail. It can be concluded that for the purpose of robotic manipulation continuous action domain algorithms are the most fruitful and applicable. Further, it can be observed that there is a trend towards exploration of sample efficient and time efficient algorithms, having solved both continuous state and action space problems. Breakthroughs in these domains will have significant impacts in the field of robotics learning.\\
 Also, as demonstrated from current state of the art in DRL, the approaches fail to handle complex policies. A reason could be that complicated policies require more samples to learn and even a sophisticated reward function. This observation highlights a void in RL in robotics. There is a need to learn highly complicated reward functions and methods to represent highly skilled behaviours and skills. This are of Inverse reinforcement learning needs to be paid more attention while learning policies using DRL. After all, complexity of the reward function is proportional to the policy complexity.\\
 Reinforcement learning is an evolved form of the cognitive architecture SORE. There is a need to reconnect the new DRL approaches to its roots in cognitive science. The problems in DRL might find extremely useful insights from theories and empirical evidence from cognitive psychology.\\
 One of the important drawbacks of DRL algorithms and visuo-motor architectures are the lack of capability of transfer learning. It is difficult to transfer skills and use the knowledge of already learnt policies to learn even complicated policies. A mechanism needs to be developed so that policies doesn't have to be learnt from scratch, but can be inherited.\\
 Many problems with temporally spread out rewards lead to credit assignment problem in RL. Thus, the reward structure too needs to be redesigned. There have been several works in incorporating intrinsic motivation in reinforcement learning as a method to induce temporal abstractions in agents \cite{barto}. These setups known as semi-Markovian decision process can be used to learn hierarchical planning actions by learning step by step about the task at hand, just as a human does.\\
 Another most important aspect of DRL that hasn't been touched upon in the main body is an approach known as guided policy search (GPS). This is because of it incipient stages in DRL currently, but the approach holds significant potential in learning robotic tasks with minimal trials. The central idea behind the algorithm is to mix model based and model free algorithms and use linear models to generate samples in order to guide the learning process. This seems like a valid assumption as humans/animals don't always learn actions from scratch, but take advantage of already well developed models of their body and physics.

\end{document}